\documentclass{article}
\usepackage{spconf,amsmath,graphicx}

\usepackage{amssymb} % define this before the line numbering.
\usepackage{color}
\usepackage{algorithm}
\usepackage{algorithmic}
\usepackage{tabu}
\usepackage{tabularx}
\usepackage{subfigure}
\usepackage{multirow}

\usepackage{verbatim}

\let\OLDthebibliography\thebibliography
\renewcommand\thebibliography[1]{
	\OLDthebibliography{#1}
	\setlength{\parskip}{0pt}
	\setlength{\itemsep}{0pt plus 0.3ex}
}

\usepackage{caption}
\title{Joint Learning of Self-representation and Indicator for Multi-view Image Clustering}
\name{Songsong Wu$^{1,5}$, Zhiqiang Lu$^1$, Hao Tang$^2$, Yan Yan$^3$, Songhao Zhu$^1$, Xiao-Yuan Jing$^4$, Zuoyong Li$^5$
}
\address{$^1$Nanjing University of Posts and Telecommunications \, $^2$University of Trento \, \\
    $^3$Texas State University \, $^4$ Wuhan University \, \\ $^5$ Industrial Robot Application of Fujian University Engineering Research Center Minjiang University
}
\begin{document}
%\ninept
%
\maketitle
\begin{abstract}
Multi-view subspace clustering aims to divide a set of multisource data into several groups according to their underlying
subspace structure. Although the spectral clustering based
methods achieve promotion in multi-view clustering, their
utility is limited by the separate learning manner in which
affinity matrix construction and cluster indicator estimation
are isolated. In this paper, we propose to jointly learn the
self-representation, continue and discrete cluster indicators
in an unified model. Our model can explore the subspace
structure of each view and fusion them to facilitate clustering simultaneously. Experimental results on two benchmark
datasets demonstrate that our method outperforms other existing competitive multi-view clustering methods.
\end{abstract}
\section{Introduction}
\label{sec:intro}
In real-world applications, images always own multiple views as they are represented by multiple ways. For instance, three typical views are pixel intensity, LBP feature, and Gabor coefficients. Unsupervised clustering of images based on their multiple views has attracted increasing attention recently \cite{c32,c33,c34}. Compared with single-view clustering (SVC) \cite{c7,tang2019fast, c6,c11,c12,c16, Tang_IJCAI16} that performs clustering relying on a single type of feature, multi-view clustering (MVC) exploits the complementary information among views to facilitate clustering.

Multi-view clustering can be divided into three categories, co-training MVC \cite{c5,c8,c9}, multiple kernel learning MVC \cite{c10,c31}, and subspace learning MVC \cite{c7,c17,c18,c19,c22,c25,c27}. Co-training based methods pursue the optimal clustering in each view while minimize the disagreement among them simultaneously. Multiple kernel learning based  methods claims that the distinct information from each view can be integrated with a linear combination of multiple kernels to promote clustering performance. Subspace learning based methods model data views as distinct low-dimensional subspaces. In this paper, we focus on subspace MVC because of it's simplicity in computation and stability in performance.

Subspace learning based MVC clustering methods rely on affinity matrix learning followed by cluster indicator prediction. However, these two procedures are accomplished separately, resulting that they are hardly optimal for clustering. Besides, the desired discrete clustering indicator is approximated by a continuous matrix for the convenience in model solving. This incurs the loss in cluster information, as reported by \cite{c3,c26} for single view clustering.

To address the above issues, we proposed a joint learning model for MVC in this paper. It can jointly learn subspace representations, continuous labels, and discrete labels from the multi-view data. In solving the optimization problem, continuous clustering label is only used as intermediate product. The main contributions of our work is to combine subspace learning, continuous label learning, and discrete label learning to obtained multi-view affinity matrices constantly updated with the inherent interactions between three steps.

%Do not use any additional Latex macros.

%-------------------------------------------------------------------------

\section{Related work}
SPC \cite{c3} selects the most informative to perform standard spectral clustering algorithm.
SSC \cite{c6} aims to find a sparse representation from the data, then the final clustering result can be obtained by spectral clustering.
S3C \cite{c16} executes clustering on each view and selects the best performance.
Min-Dis \cite{c5} aims to minimize the disagreement of two-view by generating a bipartite graph and uses spectral clustering to obtain the clustering results.
ConvexReg SPC \cite{c15} learns a common representation for all views, and executes the standard spectral clustering to get clustering results.
RMSC \cite{c14} is robust multi-view spectral clustering method and uses a Markov chain to cluster.
Di-MSC \cite{c17} adopts the HSIC criterion norm to obtain the diverse representations of different views with considering the complementary information, then uses spectral clustering to generate the clustering result.
LT-MSC \cite{c18} obtains the final clustering result by capturing high-dimension information of multi-view data.
ECMSC \cite{c25} considers complementary information with exclusivity term and guarantees the consistency with a common indicator and uses the spectral clustering algorithm to obtain clustering results.
CSMSC \cite{c27} tries to explore a common representation and specific representation from multi-view data, and the clustering result can be obtained by spectral clustering.

\begin{comment}
\cite{c19} integrates subspace learning and spectral clustering into an unified optimization framework. \cite{c25} exploits the complementary information among different views by learning an exclusive representation. A wildly spread strategy used by current multi-view clustering methods includes three parts, representation learning via subspace learning in each view, common affinity matrix construction by self-representation and spectral clustering based on the affinity matrix. For example, \cite{c17} adopts HSIC criterion to find the diversity representation of the data; \cite{c18} captures high-dimensional data information by low-rank tensor;the method \cite{c24} is to find the latent representation of each view;\cite{c27} tries to discover a common representation and specific representations from multi-view data.
\end{comment}

Beyond the above methods estimate discrete cluster indicators with continuous indicators, \cite{c23,c26} learn discrete label and then use discrete cluster labels directly for clustering in single-view. s for as we have known, the issue of discrete and continuous cluster indicator has not been touched in the case of multi-view clustering. We propose to simultaneously learn subspace representation, continuous indicator and discrete indicator for multi-view clustering, rather than straightly extend \cite{c23,c26} from single view clustering to multiple view clustering.

\section{The proposed method}
%----------------------------------------------------------------------
\subsection{Formulation}
Given sample set $X_v\in R^{d_v\times N}$,$v\in[1,2,...V]$ with $V$ views, where $N$ is the number of samples, and $d_v$ is the sample dimension of view $v$. The self-representation of samples for a single view is written as
\begin{equation}
X_v=X_vZ_v+E_v,\quad \mbox{s.t.} \;\; Z_{v(i,i)}=0
\label{eq:X}
\end{equation}
where $Z_v\in R^{n\times n}$ is self-representation matrix and $E_v\in R^{d_v\times n}$ is error term.
%Then we obtain the affinity matrix of view $v$ as
%\begin{equation}
%W_v=\frac{\mid Z_v\mid+{\mid Z_v\mid}^T }{2}
%\end{equation}
%where $|\cdot|$ denotes absolute operator.

Under the principle of spectral clustering, the discrete indicator is obtained by
%then Laplacian matrix :$L_v=D_v-W_v$, which $L_v\in R^{n\times n}$,$D_v\in R^{n\times n}$ is a diagonal matrix whose diagonal elements are composed in $D_{v_{ij}}=\sum_j W_{v_{ij}}$. The clustering result is generated by spectral clustering [3].
%-----------------------------------------------------------------------
%\subsection{Our method}
%Spectral clustering adopts the affinity matrix $W_v\in R^{n\times n}$ as an input. The subspace representation $Z_v\in R^{n\times n}$ of each view usually is obtained from the data self-representation. Assuming that there are c groups in the data $X_v$, spectral clustering can solve the following problem:
\begin{equation}
\mathop{\arg\;\min_{F}\;\;\;tr(F^TL_vF)},\quad\mbox{s.t.}\;\;F\in I_{dx}
\label{eq:F}
\end{equation}
where $I_{dx}$ means that $F=[f_1,f_2,\cdots f_n]\in R^{K\times N}$ is the discrete cluster indicator matrix with each column $f_i\in{\{0,1\}}^{K\times 1}$ contains only one element $1$ indicating the cluster the sample is assigned to. The $L$ in (\ref{eq:F}) is graph Laplacian matrix with the affinity matrix $W_v=\frac{\mid Z_v\mid+{\mid Z_v\mid}^T }{2}$. In practice, the following model is solved instead as model \ref{eq:F} is NP-hard \begin{equation}
\mathop{\arg\;\min_{P}\;\;\;tr(P^{T}L_{v}P)},\quad\mbox{s.t.}\;\;P^{T}P=I
\label{eq:P}
\end{equation}
where $P\in R^{K\times N}$ is the relaxed continuous indicator, and the orthogonal constraint is to prevent from trivial solutions.

Take formula (\ref{eq:X})(\ref{eq:F})(\ref{eq:P}) into consideration, we have the following model for multi-view subspace clustering
%---------------------
\begin{comment}
\begin{equation}
\begin{split}
\mathop {\min }\limits_{{Z_v},{E_v},F,P,Q} \sum\limits_{v = 1}^V {||{X_v} - {X_v}{Z_v} - {E_v}||_F^2}  + \\{\lambda _1}\sum\limits_{v = 1}^V {||{E_v}|{|_1}
\mbox{+} {\lambda _2}\sum\limits_{v = 1}^V {Tr({P^T}{L_v}P) + } } {\lambda _{\rm{3}}}{\rm{||F - PQ||}}_F^2
\end{split}
\end{equation}
\begin{equation*}
\mbox{s.t. }Z_v(i,i)=0, P^TP=I, Q^TQ=I, F\in I_{dx}
\label{eq:model}
\end{equation*}
\end{comment}
%--------------------------------------------------

\begin{equation}
	\label{eq:model}
	\begin{aligned}
		\mathop {\arg \; \min }\limits_{{Z_v},{E_v},F,P,Q} \; & \sum\limits_{v = 1}^V  \left \{
		\begin{aligned}
			  & \|{X_v} - {X_v}{Z_v} - {E_v} \|_{F}^{2} + {\lambda_1} \|E_v \|_{1}  \\
			+ & {\lambda_2} Tr({P^T}{L_v}P)
			+ {\lambda_3} \|F - PQ\|_{F}^{2} \\
		\end{aligned}
		\right \} \\
\, \mbox{s.t.}\; & Z_v(i,i)=0, P^TP=I, Q^TQ=I, F\in I_{dx}
	\end{aligned}
\end{equation}
%----------------------(5)
where $\lambda_1$,$\lambda_2$ and $\lambda_3$ are positive hyper-parameters. The first and second term are associated with the self-representation in each view. The third term is related to continuous indicator seeking so as to avoid NP-hard issue in computation. The fourth term build a linear relation between the continuous indicator the desired discrete indicator with matrix $Q\in R^{c\times c}$ to ensure these two indicators are consistent with each other.
%--------------------------------------------------------------------------------

\subsection{Optimization}
We solve model (\ref{eq:model}) by alternative optimization strategy, i.e. optimizing one variable by fixing the previous values of other variables. To facilitate the narrative, we omit the subscript of view tentatively.

\noindent\textbf{Update $Z$.}
When $E$,$F$, $P$, and $Q$ are fixed, we have
%****************************************
\begin{equation}
\label{eq:opt_Z}
	\begin{aligned}
\mathop {\arg \;\; \min }\limits_{Z} \quad & ||X-XZ-E||_F^2 + {\lambda _2}Tr({P^T}LP) \\
		\text{s.t.}  \quad  & \; Z_{ii}=0	
	\end{aligned}
\end{equation}
%****************************************
By defining $Y_{ij}=\|P_{i}-P_{j}\|_2^2$ and using the definition of $L$, we solve (\ref{eq:opt_Z}) by seek a optimal column of $Z$ at one time based on the following optimal model
%------------------------------------------------------------------
\begin{equation}
\begin{split}
\mathop {\arg \; \min }\limits_{Z_{i}} ||{Z_{i}} - V_{i}||_2^2 + \frac{{{\lambda _2}}}{2}|{Z_{i}}{|^T}Y_{i}, \quad \text{s.t.} \; Z_{ii}=0
\end{split}
\label{eq:sz1}
\end{equation}
where $Z_{i}$, $Y_{i}$ and $X_{i}$ are the $i$-th column of $Z$,$Y$, and $X$ respectively, and $V_{i}=\frac{K^TX_i}{X_i^TX_i}$ with $K=X-(XZ-X_{i}Z_{i}^T)-E$. So far, we can get closed form solution based on (\ref{eq:sz1}) for that if $j=i$, we have $Z_{ji}=0$, and if $j\neq i$, we have
\begin{equation}
\begin{array}{l l}
Z_{ji} = & sign(V_{ji})(|V_{ji}| - \frac{\lambda_2Y_{ji}}{4}) +  \\
&  \left\{
\begin{array}{l l}
    V_{ji} - \frac{\lambda_{2}Y_{ji}}{4}, & \text{if} \;\; V_{ji} >  + \frac{\lambda_2Y_{ji}}{4}\\
    V_{ji} + \frac{\lambda_{2}Y_{ji}}{4}, & \text{if} \;\;  V_{ji} <  - \frac{\lambda_2Y_{ji}}{4}\\
    0, & otherwise
\end{array} \right.
\end{array}
\label{eq:solution_Z}
\end{equation}
%------------------(13)

\noindent\textbf{Update $E$.} When $Z$,$F$,$P$, and $Q$ are fixed, we seek the optimal $E$ using the same column-wise strategy as $Z$. Specifically, for a single column $E_{i}$ of $E$, we have
\begin{equation}
\mathop {\arg \; \min }\limits_{{E_i}} ||{(X - XZ)_i} - {E_i}||_F^2 + {\lambda _1}{\rm{||}}{E_i}{\rm{|}}{{\rm{|}}_{\rm{1}}}
\end{equation}
Where $(X-XZ)_i$ is the $i$-th column in the matrix. The solution is provided by
\begin{equation}
\begin{split}
E_{ji} = sign(X - XZ)_{ji}(|(X - XZ)_{ji}| - \frac{{\lambda_1}}{2}) \\+ \left\{ \begin{array}{l}
(X - XZ){ji} - \frac{{{\lambda _1}}}{2}, if \ (X - XZ)_{ji} > \frac{{{\lambda _1}}}{2}\\
(X - XZ){ji} + \frac{{{\lambda _1}}}{2}, if \ (X - XZ)_{ji} < \frac{{{\lambda _1}}}{2}\\
0, otherwise
\end{array} \right.
\end{split}
\label{eq:solution_E}
\end{equation}
%--------------(16)

\noindent\textbf{Update P.}
When Z, F, E, and Q are fixed, the model becomes
\begin{equation}
\begin{split}
\mathop {\arg \min }\limits_P {\lambda _{\rm{2}}}Tr({P^T}LP){\rm{ + }}{\lambda _{\rm{3}}}{\rm{||}}F{-}PQ{\rm{||}}_F^2, \; \text{s.t.} P^TP{=}I
\end{split}
\label{eq:solution_P}
\end{equation}
%----------(17)
The model can be efficiently solved by the algorithm proposed by \cite{c13}.

\noindent\textbf{Update Q.}
When F, Z, E and P are fixed, the model becomes
\begin{equation}
\begin{split}
\mathop {\arg\;\min }\limits_Q {\lambda _{\rm{3}}}{\rm{||}}F{-} PQ{\rm{||}}_F^2, \quad \text{s.t.} \; Q^TQ=I
\end{split}
\end{equation}
%-----------(18)
It is a orthogonal Procrustes problem [28], whose solution is given as
\begin{equation}
Q=UV^T
\label{eq:solution_Q}
\end{equation}
where $U$ and $V$ are the left and right singular value matrices of $F^TP$.

\noindent\textbf{Update F.}
When Z, E, P and Q are fixed, the model becomes
\begin{equation}
\begin{split}
\mathop {\arg \; \max }\limits_F \;\; {\lambda _{\rm{3}}}Tr({F^T}PQ), \quad \text{s.t.} \; F\in I_{dx}
\end{split}
\end{equation}
%----------(20)
whose optimal solution can be obtained as follows
\begin{equation}
\begin{split}
{F_{ij}} = \left\{ \begin{array}{l}
1, if \; j = \arg {\max \limits_k}{(PQ)_{ik}}\\
0,otherwise
\end{array} \right.
\end{split}
\label{eq:solution_F}
\end{equation}
%---------(22)
In summary, the proposed method can be expressed in Algorithm~\ref{alg}.

\renewcommand{\algorithmicrequire}{ \textbf{Input:}} %Use Input in the format of Algorithm
\renewcommand{\algorithmicensure}{ \textbf{Output:}} %UseOutput in the format of Algorithm
\begin{algorithm} [!t]
\caption{The proposed MVC method}
\begin{algorithmic}
\label{alg}
\REQUIRE~$V$ view image sets $\{X=_v\}_{v=1}^{V}$, positive parameters $\lambda_1$,$\lambda_2$,$\lambda_3$.

Initialize $Z_1,Z_2,\cdots,Z_V,P,Q$ with random values, and set $F,E=0$
\REPEAT
\STATE 1.Update $Z_v$ with (\ref{eq:solution_Z})

\STATE 2.Update $E_v$ with (\ref{eq:solution_E})

\STATE 3.Update $P$ with (\ref{eq:solution_P}).

\STATE 4.Update $Q$ with (\ref{eq:solution_Q})

\STATE 5.Update $F$ with (\ref{eq:solution_F})

\UNTIL{Stopping criterion}

\STATE Compute affinity matrix by by $W=\sum\limits_{v = 1}^V \frac{\mid Z_v\mid+{\mid Z_v\mid}^T }{2}$

\STATE Perform spectral clustering using affinity matrix $W$.

\ENSURE~Clustering result.
\end{algorithmic}
\end{algorithm}

%===========================================================
% \begin{table*} \small
% \begin{center}
% \large
% \begin{tabu} to 1\textwidth{X[2,c]|X[2,c]|X[2,c]|X[2,c]}
% \hline
% Dataset &Views &Clusters &Samples\\
% \hline
% ORL &3 &40 &400\\
%  \hline
% Yale &3 &15 &165  \\
% \hline
% \end{tabu}
% \end{center}
% \caption{Statistics of the multi-view datasets}
% \end{table*}

\section{Experiment}
\subsection{Experimental Settings}
\noindent\textbf{Dataset Description.} We adopted two face image datasets in the experiments which are used widely in image clustering [17, 18, 25]. In our experiment, each image sample has three views: intensity, LBP \cite{c4}, and Gabor \cite{c1}. Statistics of the datasets are summarized in Table 1.
(i) ORL is divided into 40 different themes, each containing 10 images. All images are in a uniform black background and are All images are taken from the top of the front. For some of these themes, the images are taken at different times. In light conditions, facial expressions (such as open eyes, closed eyes, laugh, and non-laugh), facial details (glasses) and other aspects are different.
(ii) Yale is consisted of 165 gray-scale images of 15 individuals, each individual has 11 images, including: center light, glasses, happy, left light, no glasses, normal, right light, sad, sleepy, surprised and wink.

%%%%%%%%%%%%%%%%%
\iffalse
\begin{table}[!t]
	\centering
	\caption{Statistics of the multi-view datasets.}
	\begin{tabular}{lccc} \hline
	Dataset &Views & Clusters &Samples \\ \hline
	ORL &3 &40 &400 \\ \hline
    Yale &3 &15 &165 \\ \hline
	\end{tabular}
	\label{tab:amt}
	\vspace{-0.6cm}
\end{table}
\fi
%%%%%%%%%%%%%%%%%

\noindent\textbf{Comparison Algorithm.}
We compare our method with three single-view methods (SPC \cite{c3}, SSC \cite{c6}, S3C \cite{c16}) and seven multi-view methods (Min-Dis \cite{c5}, ConvexReg SPC \cite{c15}, RMSC \cite{c14}, Di-MSC \cite{c17}, LT-MSC \cite{c18}, ECMSC \cite{c25}, CSMSC \cite{c27}). The optimal parameters of the proposed method are set empirically based on grid searching. Specifically, for ORL dataset, the parameters are set as:  $\lambda_1{=}0.002$,$\lambda_2{=}0.5$, $\lambda_3{=}0.1$ and for Yale dataset, they are set as: $\lambda_1{=}0.003$, $\lambda_2{=}0.7$, $\lambda_3{=}0.2$. The parameter values of other methods are set according to the corresponding papers.

\begin{table*}[!t] \small
\begin{center}
\caption{Clustering performances on ORL dataset (mean±standard deviation).}
\scriptsize
\begin{tabu} to 1\textwidth{|X[2,c]|X[2,c]|X[2,c]|X[2,c]|X[2,c]|X[2,c]|X[2,c]|X[2,c]|}
\hline
\quad &Method	&NMI	&ACC	&ARI	&F	&P	&Re\\
\hline
single &SPCbest	&0.884(0.002)	&0.726(0.025)	&0.655(0.005)	&0.664(0.005)	&0.610(0.006)	&0.728(0.005)\\
\hline
single &SSCbest	&0.893(0.007)	&0.765(0.008)	&0.694(0.013)	&0.682(0.012)	&0.673(0.007)	&0.764(0.005)\\
\hline
single &S3Cbest	&0.902(0.012)	&0.784(0.009)	&0.705(0.019)	&0.698(0.018)	&0.688(0.012)	&0.791(0.011)\\
%\hline
%Multiple &FeaCon &0.835(0.004)	&0.675(0.028)	&0.564(0.010)	&0.574(0.010)	&0.532(0.011)	&0.624(0.008)\\
\hline
Multiple &Min-Dis	&0.876(0.002)	&0.748(0.051)	&0.654(0.004)	&0.663(0.004)	&0.615(0.004)	&0.718(0.003)\\
\hline
Multiple &RMSC	&0.872(0.012)	&0.723(0.025)	&0.644(0.029)	&0.654(0.028)	&0.607(0.033)	&0.709(0.027)\\
\hline
Multiple &ConReg	&0.883(0.003)	&0.734(0.031)	&0.668(0.032)	&0.676(0.035)	&0.628(0.041)	&0.731(0.030)\\
\hline
Multiple &LTMSC	&0.930(0.002)	&0.795(0.007)	&0.750(0.003)	&0.768(0.007)	&0.766(0.009)	&0.837(0.004)\\
\hline
Multiple &DiMSC	&0.940(0.003)	&0.838(0.001)	&0.802(0.000)	&0.807(0.003)	&0.764(0.012)	&0.856(0.004)\\
\hline
Multiple &ECMSC	&\textcolor{red}{0.947(0.009)}	&0.854(0.011)	&0.810(0.012)	&0.821(0.015)	&0.783(0.008)	&0.859(0.012)\\
\hline
Multiple &CSMSC	&0.942(0.005)	&0.868(0.012)	&0.827(0.002)	&0.831(0.001)	&\textcolor{red}{0.860(0.002)}	&0.804(0.003)\\
\hline
Proposed & Ours &0.943(0.005)	&\textcolor{red}{0.886(0.016)}	&\textcolor{red}{0.831(0.019)}	&\textcolor{red}{0.835(0.019)}	&0.804(0.023)	&\textcolor{red}{0.868(0.018)}\\
\hline
\end{tabu}
\end{center}
\vspace{-0.6cm}
\label{table:orl}
\end{table*}

\begin{table*}[!t]\small
\label{table:yale}
\begin{center}
\caption{Clustering performances on ORL dataset (mean±standard deviation).}
\scriptsize
\begin{tabu} to 1\textwidth{|X[2,c]|X[2,c]|X[2,c]|X[2,c]|X[2,c]|X[2,c]|X[2,c]|X[2,c]|}
\hline
\quad &Method	&NMI	&ACC	&ARI	&F	&P	&Re\\
\hline
single &SPCbest	&0.654(0.009)	&0.616(0.030)	&0.440(0.011)	&0.475(0.011)	&0.457(0.011)	&0.495(0.010)\\
\hline
single &SSCbest	&0.671(0.011)	&0.627(0.000)	&0.475(0.004)	&0.517(0.007)	&0.509(0.003)	&0.547(0.004)\\
\hline
single &S3Cbest	&0.678(0.013)	&0.634(0.016)	&0.471(0.005)	&0.508(0.012)	&0.512(0.005)	&0.568(0.025)\\
%\hline
%Multiple &FeaCon	&0.665(0.037)	&0.578(0.038)	&0.396(0.011)	&0.434(0.011)	&0.419(0.012)	&0.450(0.009)\\
\hline
Multiple &Min-Dis	&0.645(0.005)	&0.615(0.043)	&0.433(0.006)	&0.470(0.006)	&0.446(0.005)	&0.496(0.006)\\
\hline
Multiple &RMSC	  &0.684(0.033)	  &0.642(0.036)	 &0.485(0.046)	&0.517(0.043)	&0.500(0.043)	&0.535(0.044)\\
\hline
Multiple &ConReg	&0.673(0.023)	&0.611(0.035)	&0.466(0.032)	&0.501(0.030)	&0.476(0.032)	&0.532(0.029)\\
\hline
Multiple &LTMSC	&0.765(0.008)	&0.741(0.002)	&0.570(0.004)	&0.598(0.006)	&0.569(0.004)	&0.629(0.005)\\
\hline
Multiple &DiMSC	 &0.727(0.010)	&0.709(0.003)	&0.535(0.001)	&0.564(0.002)	&0.543(0.001)	&0.586(0.003)\\
\hline
Multiple &ECMSC	 &0.773(0.010)	&0.771(0.014)	&0.590(0.014)	&0.617(0.012)	&0.584(0.013)	&0.653(0.013)\\
\hline
Multiple &CSMSC	 &\textcolor{red}{0.784(0.001)}	&0.752(0.001)	&0.615(0.005)	&0.640(0.004)	&\textcolor{red}{0.673(0.002)}	&0.610(0.006)\\
\hline
Proposed &Ours	 &0.782(0.005)	&\textcolor{red}{0.792(0.026)}	&\textcolor{red}{0.620(0.008)}	&\textcolor{red}{0.644(0.007)}	&0.616(0.009)	&\textcolor{red}{0.661(0.006)}\\ \hline
\end{tabu}
\end{center}
\vspace{-0.6cm}
\end{table*}

\noindent\textbf{Evaluation Metrics.}
In our experiment, we use six metrics to evaluate the clustering performance, i.e., Normalized Mutual Information (NMI), Accuracy (ACC), Adjusted Rand Index (ARI), F-score, Precision and Recall. ACC and NMI have been adopted to evaluate clustering result \cite{c20,c21}. Precision, F-score, ARI and Recall have been widely used in \cite{c29} to measure the clustering quality. These six metrics evaluate the different performance of clustering. For all metrics, higher values indicate better cluster quality.
%--------------------------------------------------------------

\noindent\textbf{Performance Comparison.}
We repeat the random experiments for $30$ rounds and report the mean and standard deviation as final results. The  clustering results on ORL and Yale are provided in Table 1 and Table 2 respectively.

%Table 1 shows the performance comparison on the ORL dataset.
From Table 1, we observe that compared with Di-MSC, we achieve improvements around $0.3\%$, $4.8\%$, $2.9\%$, $2.8\%$, $4.0\%$, and $1.2\%$ in terms of NMI, ACC, ARI, F, P, and Re. Compared with ECMSC, we achieve improvements around $3.2\%$, $2.1\%$, $1.4\%$, $2.0\%$, and $0.9\%$ in terms of ACC, ARI, F, P, and Re, because the affinity matrix is constantly updated due to add learning about discrete labels. Compared with latest method (CSMSC), our experimental results improve around $1.8\%$, $0.4\%$, $0.4\%$, and $6.4\%$ in terms of ACC, ARI, F, Re. The reason is that we consider the relationship between subspace learning and spectral clustering. But our results are lower than CSMSC in NMI and P, it seems that we did not consider the complementary information of multiple views.

\begin{figure} [!t] \small
\centering
\subfigure{
%\label{Fig.sub.1}
\includegraphics[width=4cm, height=3cm]{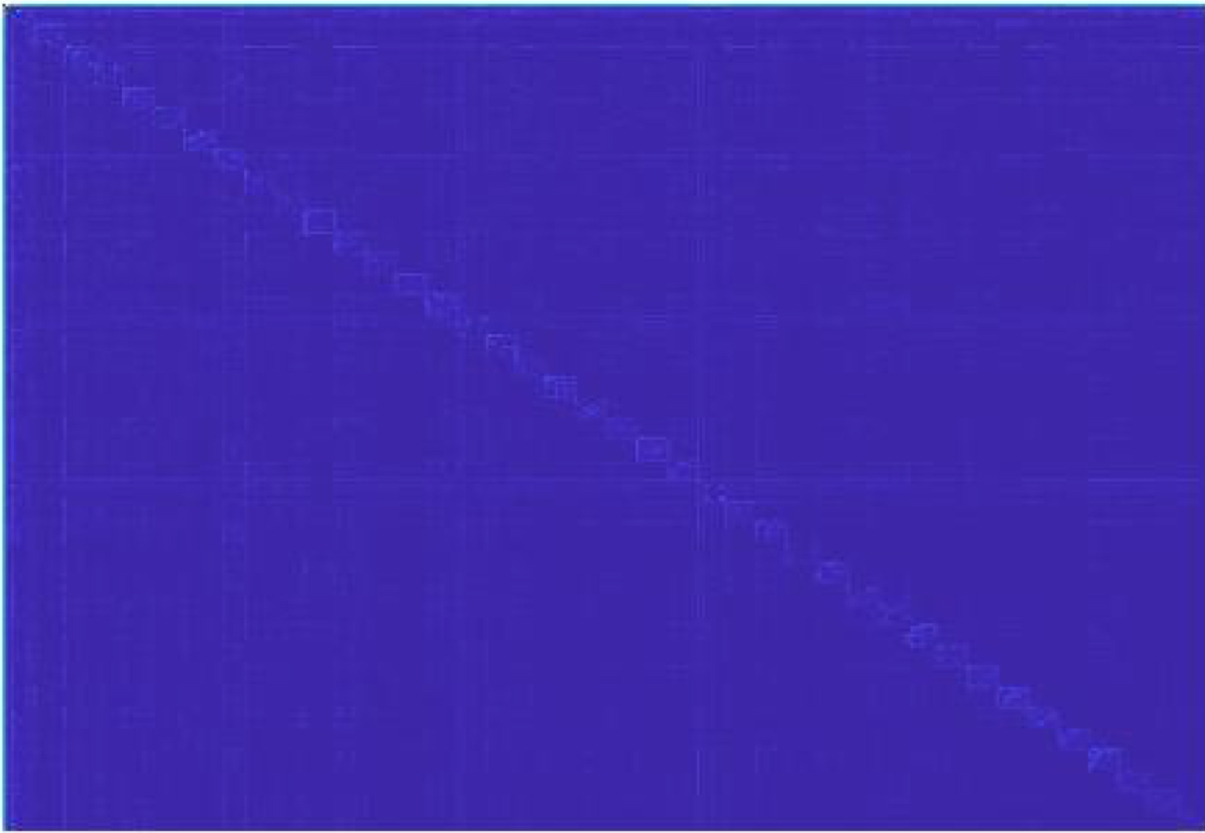}}
\subfigure{
%\label{Fig.sub.2}
\includegraphics[width=4cm, height=3cm]{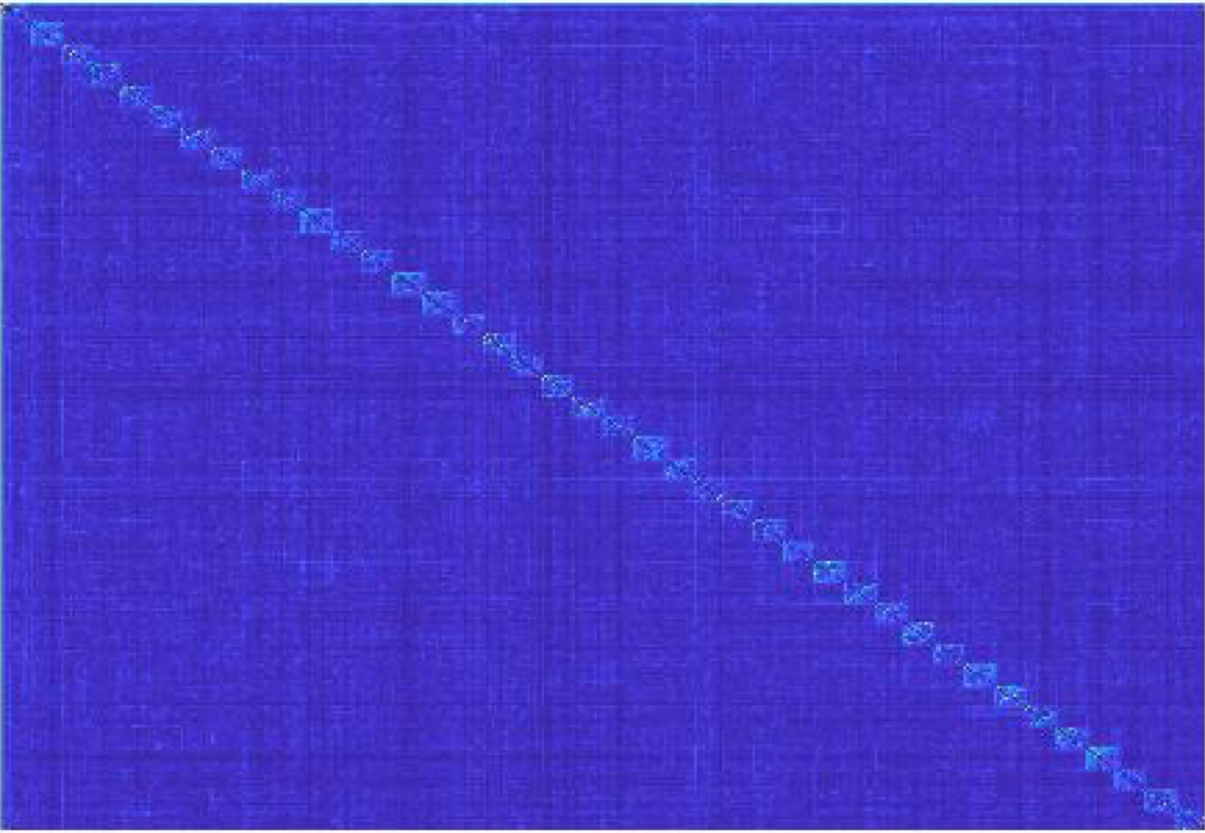}}
\subfigure{
%\label{Fig.sub.3}
\includegraphics[width=4cm, height=3cm]{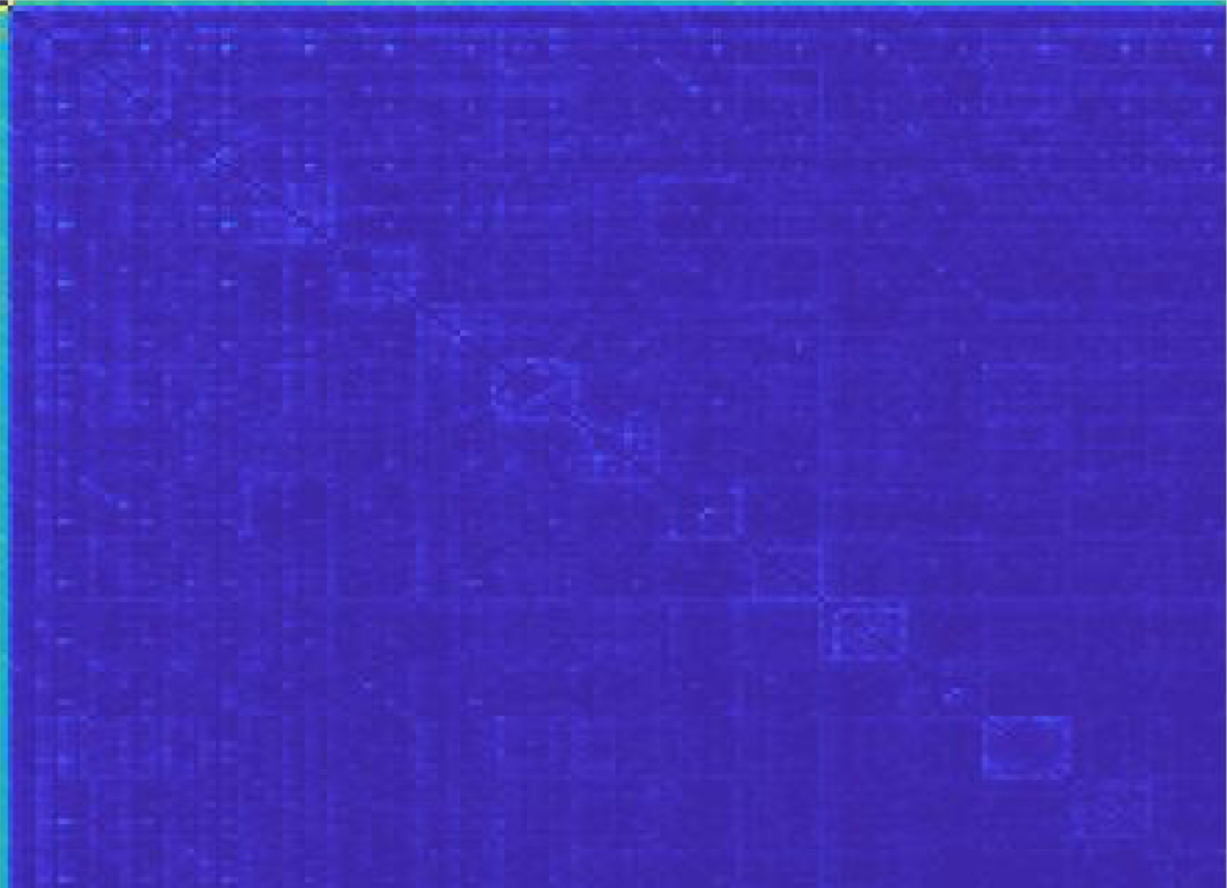}}
\subfigure{
%\label{Fig.sub.4}
\includegraphics[width=4cm, height=3cm]{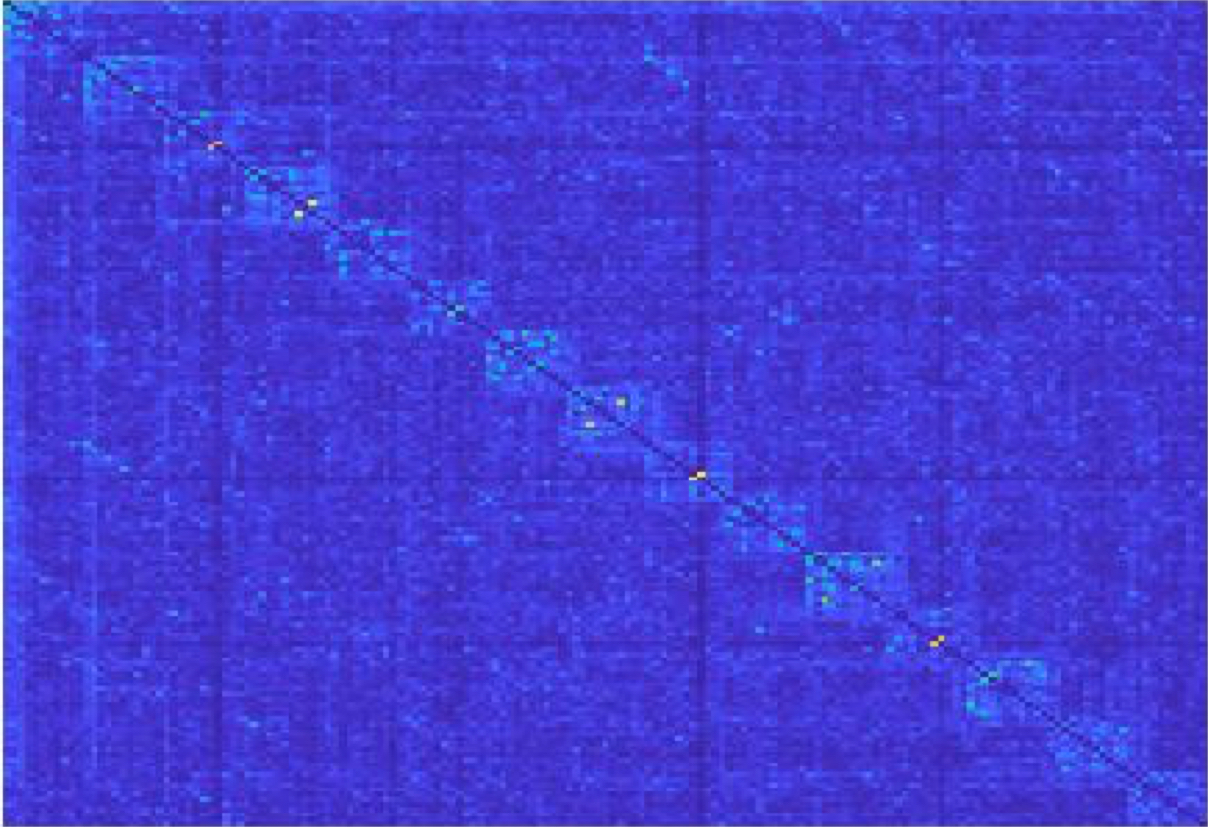}}
\caption{Affinity matrix visualization on ORL (top) and Yale
(bottom). From left to right: The affinity matrix without joint
learning (corresponding to $\lambda_2=0, \lambda_3=0$) and the affinity
matrix of our method.}
\label{fig:visualiation}
\vspace{-0.6cm}
\end{figure}

Table 2 shows that the multi-view clustering results on Yale dataset. The method LT-MSC has a significant improvement. It seems that low-rank tensor is suitable for images clustering. Di-MSC considers the diversity representations of different views. Compared with Di-MSC, we achieve improvements around $5.5\%$, $8.3\%$, $9.5\%$, $8.0\%$, $7.3\%$, $7.5\%$ in terms of NMI, ACC, ARI, F, P, Re, respectively. The major reason is that we put subspace representation and spectral clustering into an optimization model. Compared with ECMSC, we achieve improvements around $0.9\%$, $2.1\%$, $3.0\%$, $3.7\%$, $3.2\%$, $0.8\%$ in six metrics, respectively. The reason is that our method does not use continuous label to update the subspace representation, instead simultaneously uses discrete label and continuous label to update the subspace representation. Compared with latest method (CSMSC), our experimental results improve around $4.0\%$, $0.5\%$, $0.4\%$, $5.0\%$ in terms of ACC, ARI, F, and Re. CSMSC does not put subspace learning and spectral clustering method in an optimization model.

\begin{figure}[!t]
\centering
\subfigure[ORL]{
\label{Fig.sub.1}
\includegraphics[height=3cm]{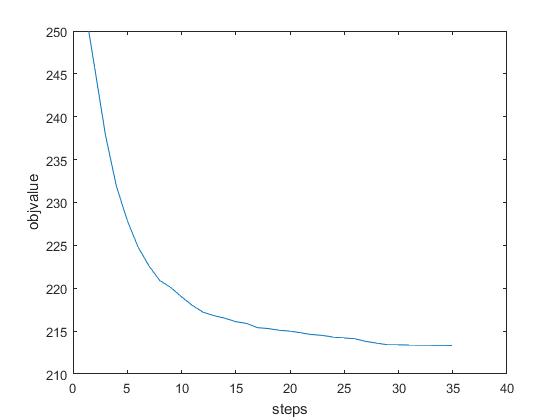}}
\subfigure[Yale]{
\label{Fig.sub.2}
\includegraphics[height=3cm]{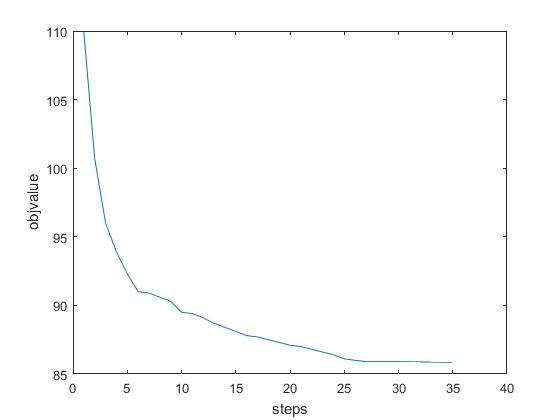}}
\caption{Convergence curve of our method on ORL and Yale.}
\label{fig:convergence}
\vspace{-0.6cm}
\end{figure}

\noindent\textbf{Visualization and  Convergence Analysis.}
To validate the effectiveness of our method, we show the visualization of affinity matrices on Yale and ORL dataset. As observed in Fig.\ref{fig:visualiation}, we can clearly see that the affinity matrix of our method presents multiple block structures. Each block structure represents a subspace. From the two figures on the right, we can see simultaneously learn subspace representations of each view, continuous labels, and discrete labels will be beneficial for multi-view subspace clustering.
As shown in Fig.~\ref{fig:convergence}, ``objvalue'' represents the value of the objective function; ``steps'' is the number of iterations. We show the convergence of our method on ORL and Yale datasets.
We can see that our algorithm converge quickly on both datasets.

%-------------------------------------------------
\section{Conclusions}
We propose a novel multi-view image clustering approach by jointly learning the self-representation of images and cluster indicators. The two processes promote each other to achieve complementary information confusion for multi-view clustering. The experimental results of multi-view face images clustering demonstate that our method outperforms other existing competitive multi-view clustering algorithms.

\section{Acknowledgement}
This work was supported in part by Industrial Robot Application of Fujian University Engineering Research Center, Minjiang University. (Grant No. MJUKF-IRA201806).

\clearpage
\small
\bibliographystyle{IEEEbib}
\bibliography{ampc_clustering}

\end{document}